\documentclass[letterpaper, 10 pt, conference]{ieeeconf}  

\IEEEoverridecommandlockouts                              

\usepackage{graphics} 
\usepackage{epsfig} 
\usepackage{amsmath} 
\usepackage{amssymb}  
\usepackage{algorithm}
\usepackage{algorithmic}

\usepackage[noadjust]{cite}
\usepackage{todonotes}
\usepackage{hyperref}
\usepackage{graphicx}
\usepackage{subcaption}
\usepackage{multirow}

\title{\LARGE \bf
Map as The Hidden Sensor: Fast Odometry-Based Global Localization
}

\author{Cheng Peng$^{1,2}$ and David Weikersdorfer$^{1}$
\thanks{*This work was supported by NVIDIA} 
\thanks{$^{1}$Cheng Peng and $^{1}$David Weikersdorfer are with ISAAC SDK team, NVIDIA Corperation,
Santa Clara, CA 95051}
\thanks{$^{2}$Cheng Peng is with the Department of Computer Science, University of Minnesota, Twin Cities, Minneapolis, MN, 55455
        {\tt\small peng0175@umn.edu}}%
}

\begin{document}

\maketitle
\thispagestyle{empty}
\pagestyle{empty}

\begin{abstract}
Accurate and robust global localization is essential to robotics applications. We propose a novel global localization method that employs the map traversability as a hidden observation. The resulting map-corrected odometry localization is able to provide an accurate belief tensor of the robot state. Our method can be used for blind robots in dark or highly reflective areas. In contrast to odometry drift in long-term, our method using only odometry and the map converges in long-term.
Our method can also be integrated with other sensors to boost the localization performance. The algorithm does not have any initial state assumption and tracks all possible robot states at all times. Therefore, our method is global and is robust in the event of ambiguous observations. We parallel each step of our algorithm such that it can be performed in real-time (up to $\sim 300$ Hz) using GPU. We validate our algorithm in different publicly available floor-plans and show that it is able to converge to the ground truth fast while being robust to ambiguities. 
\end{abstract}

\section{Introduction}

Accurate and robust self-localization is one of the fundamental tasks for indoor robots. If a robot is provided with a false location, it can be catastrophic to other core functionalities such as planning and navigation. Most localization methods integrate odometry information generated by differential wheels or inertial measurement unit (IMU). The advantage of odometry is the local consistency. The trajectory is guaranteed to be smooth, since it is built by accumulating odometry data over time. 
However, the trajectory will drift due to error accumulation. 
\begin{figure}[t!]
    \centering
   \begin{subfigure}[b]{0.48\textwidth}
       \includegraphics[width=\textwidth]{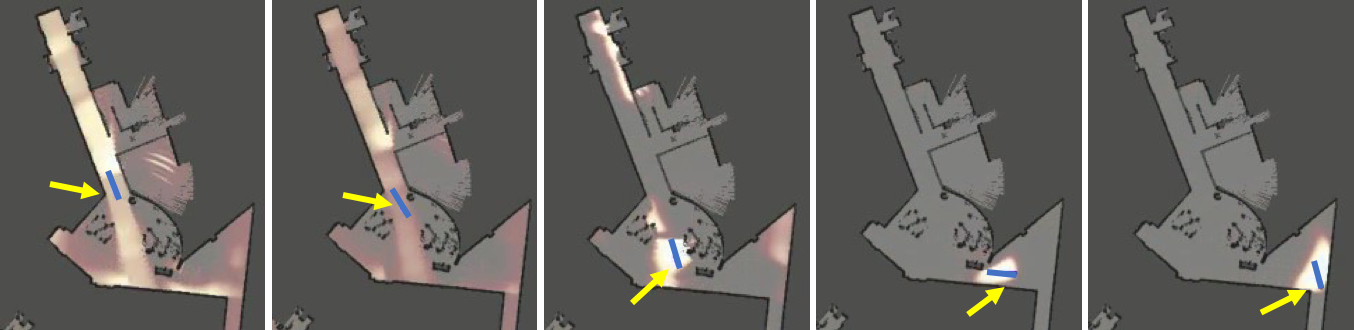}
       \vspace*{-6mm}
       \caption{Odometry + Map}
   \end{subfigure}
   \begin{subfigure}[b]{0.48\textwidth}
       \includegraphics[width=\textwidth]{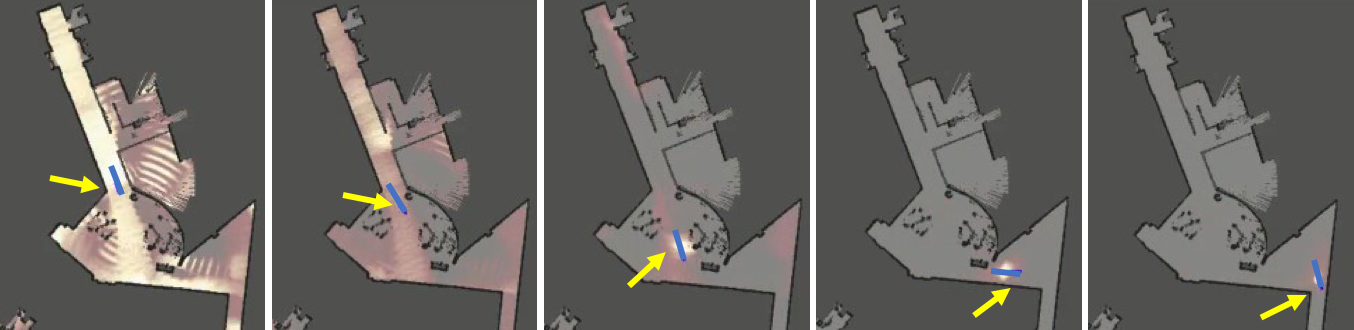}
       \vspace*{-6mm}
       \caption{Odometry + Map + Sample Updates}
    \end{subfigure}
    \begin{subfigure}[b]{0.48\textwidth}
       \includegraphics[width=\textwidth]{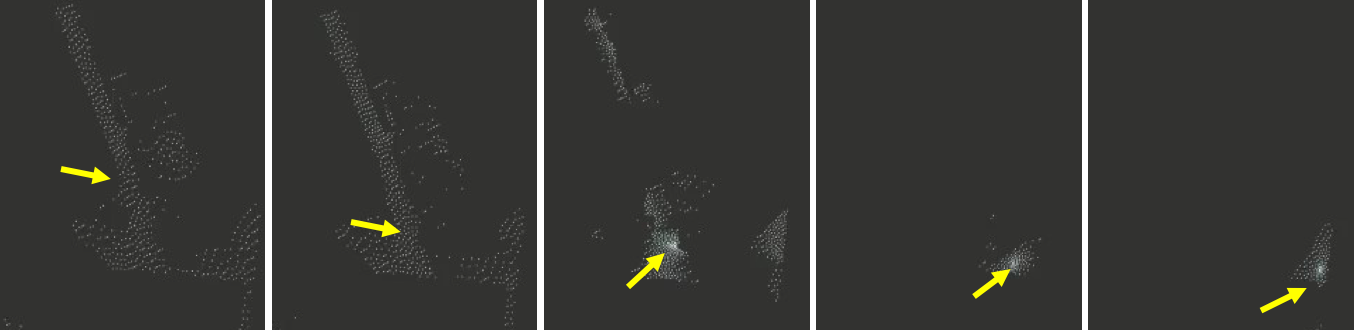}
       \vspace*{-6mm}
       \caption{Samples from Belief Map}
    \end{subfigure}
    \caption{The changes of the belief map through time (from left to right). Yellow arrows point to the ground truth in blue. (a) The belief map using only odometry with map-correction. (b) The belief map with LIDAR sample updates which converges faster than the previous version. (c) The robot state samples using the Floyd Steinberg method~\cite{floyd1976}.}
    \label{fig:teaser}
\end{figure}
To improve the localization accuracy, prior works generalize and fuse other sensors which have descriptive observations. The observations from different sensors such as visual features~\cite{GalvezTRO12, oliva2001modeling, schonberger2018semantic, zhou2016learning}, 2D/3D point clouds~\cite{yang2016fast}\cite{dube2017segmatch} from cameras, and LIDAR sensors are combined with odometry information to correct the drift and provide efficient localization and tracking~\cite{hess2016real, montemerlo2003fastslam, se2005vision, grisetti2007fast, mur2015orb, forster2014svo, dellaert1999monte}. However, such vision-based auxiliary features are not always available. For example, visual features fail in dark environments and LIDAR features fail in heavily reflective environments. 

In this paper, we revisit the fundamental question: if only the odometry data and the environment map is provided, can the robot still be able to localize itself? In another words, we study whether a blind robot can localize itself fast and robustly. 
It is observed that the trajectory integrated from the odometry may penetrate through walls and obstacles (shown in Fig~\ref{fig:traj_comp}) due to noise and drift, which is physically impossible. 
Our key conjecture is that the map can be exploited as an additional sensor, and the odometry propagation should be corrected by the traversability of the map. As a proof of concept, we focus on indoor robot localization with only odometry and a map as inputs. We further illustrate that, even though originated from the scenarios where other sensors are unavailable, our method can be fused with those sources if available to boost the performance. 

We propose a method to incrementally merge odometry data with the guidance from the map. Our method generates a global state belief \textit{tensor} in real-time that not only incorporates multiple location and orientation hypotheses but also tolerates the motion uncertainty from the odometry data. The belief tensor is a dense representation such that all possible robot states can be estimated in real-time, which has never been fully explored before. Comparing to the sampling methods~\cite{gustafsson2010particle, davidson2010application, dellaert1999monte} that are semi-global without guarantees, our dense belief tensor guarantees that the correct robot state is always updated and is within high probability region. While other auxiliary sensors such as camera or LIDAR could be misleading due to similar appearance, our belief map is able to maintain multiple hypotheses of the possible robot state to prevent a catastrophic localization error. 
Using such robust and global location prior, we can also efficiently recover from localization failure. 

Our core contributions and novelties are the followings. Firstly, we propose a complete formulation of the map-corrected odometry update using recursive Bayesian method~\cite{thrun2005probabilistic} that provides robust guarantees, which other methods can not offer. Secondly, we parallel each step of our algorithm such that it can be performed in real-time (up to $\sim 300$Hz). Lastly, we construct our algorithm such that the belief tensor can be fused with other sensor (with constraint in Eq~\ref{eq:obs_constraint}) updates using efficient sampling method to improve the localization accuracy. We test our algorithm extensively in several indoor environments. We also show that even in the existence of topological ambiguities, our algorithm is still able to converge to the exact location by paying a short traveling cost.
\section{Related Work}
\subsection{Bayesian filtering approach}
One of the fundamental methods to localize robot positions is the Bayesian filter method \cite{montemerlo2003fastslam, grisettiyz2005improving, dellaert1999monte, hess2016real}. One of the approaches is the Extended Kalman Filter (EKF) \cite{maybeck1982stochastic, dissanayake2001solution}, which assumes the observation and motion models of the robots are Gaussian distributions. They have proven to be very efficient to track the locations and merge the interactions with landmarks. However, it is unable to track multiple hypotheses for non-Gaussian posterior distribution. It normally arises when the robot is lost or ambiguous observations are made.

Another approach ~\cite{dellaert1999monte} attempts to solve the arbitrary state representation using Monte Carlo sampling. The posterior state distribution is represented as a sum of discrete samples. It is also known as the \textit{particle filter}~\cite{doucet1998sequential}. Comparing to the EKF method, each particle is a robot state with weight associated with the corresponding probability density. Instead of the analytical results obtained from EKF, particle filer focuses on numerical integration. It allows for simple implementation without any linearity or Gaussian models. Despite the advantages, efficient and representative sampling of the posterior distribution is not guaranteed~\cite{aulinas2008slam}. Sampling bias will be introduced when target and proposal distributions are largely different, which is known as the particle depletion problem~\cite{kwak2007analysis, gustafsson2010particle}. It is due to the underlining Markov model properties~\cite{grisettiyz2005improving} where the wrong location can be identified as a sink and thus the system is unable to recover from such localization failure. There are some works~\cite{grisetti2007improved, grisettiyz2005improving, thrun2005probabilistic} that propose to resample the entire state space when localization failure occurs so that the particles can re-center themselves at the correct location. However, the resampling process does not necessarily guarantee a state close to the ground truth and have to rely on localization failure detection methods~\cite{alsayed2017failure, klein2008improving}, which is not trivial itself. On the contrary, our method tracks all possible states at all time and guarantees that the correct trajectory always survives even in the event of localization failure.

Another disadvantage is that the particles are distinct and they can be re-sampled extensively with similar score, which wastes computational power. It is because their states cannot be merged and have to be represented discretely.
In comparison, our idea is similar to an old work from Burgard et. al~\cite{burgard1996estimating} and~\cite{thrun2005probabilistic} called histogram filter, which has the advantages that all states can be discretely represented as grids in the map. The state probabilities can flow through the grids, which eliminates the necessity to sample and re-sample.

\subsection{Map integration}
Other works~\cite{o2006global, beauregard2008indoor, klepal2007bayesian, davidson2010application} explore the idea of using map as additional constraints to robot motion. They are particle filter based method. A particle is considered valid only if the corresponding trajectory does not intersect the obstacle map. But their methods does not provide a complete formulation to integrate with map nor any guarantees to the localization capabilities.

Using map information for localization has another set of approaches call \textit{map matching}~\cite{romero2018icra, chen2011approximate, luo2017enhanced, ochieng2003map}. Romero et. al~\cite{romero2018icra} proposed a novel method that integrates the map information. The observation is that odometry is relatively accurate in short terms and the corresponding trajectory segment is unique in certain region of the map such as turning points. By identifying those unqiue regions through short trajectory segments, localization accuracy can be improved. In contrast, our proposed algorithm can utilize the occupancy map to correct the odometry drift at any location, which provides a much stronger prior to localization. Since we only constraint the state to be within the traversable regions, the information from certain unique turning regions can be integrated as well as general places such as straight hallways. 

Instead of representing the world as occupancy grids, another method~\cite{brubaker2015map} uses topological road maps to provide guidance to odometry for highway localization. Using the angular relationships between roads, the vehicle is able to successfully locate itself after a few unique turns. This work targets outdoor localization. However, highway road structure is very unique since the vehicle must follow the road and cannot diverge to any open space. Comparing to indoor localization, if only odometry is available, large open space will be a tremendous challenge. 
\section{Method overview}
\begin{figure}[t!]
    \centering
    \includegraphics[width=0.48\textwidth]{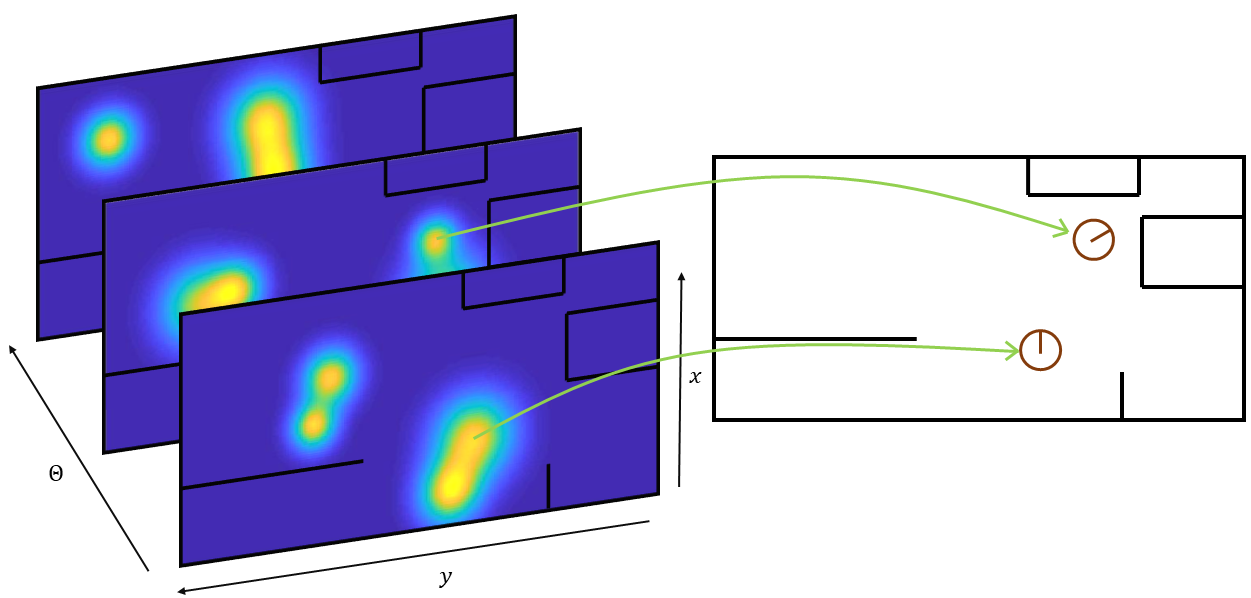}
    \caption{Illustration of the belief tensor where each channel represents different robot orientation.}
    \label{fig:heatmap}
\end{figure}
In our method, the state uncertainty distribution is updated with an occupancy map $\mathcal{M}$, where a pixel region $m$ is characterized as free $P(m_{free} \in \mathcal{M}) = 0$ or occupied space $P(m_{occ} \in \mathcal{M}) = 1$.
The state of the robot at time $t$ is $x_t \in SE(2) $ that contains both position and orientation. Each pixel will represent a state of the robot and we assume an uniform distribution for each state $\hat{x}_t$ within a predefined pixel region $q$.
\begin{equation}
    P(x_{q,t}) = |q|^{-1}\int_{q} P(\hat{x}_t) d\hat{x}_t
\end{equation}
where $|q|$ is the area of the region $q$.
The odometry motion update at time $t$ is defined as $u_t$ such that the transitional probability from state $x_{t-1}$ to $x_t$ is given as $P(x_t | u_t, x_{t-1})$. A observation from an auxiliary sensor is defined as $z_t$ and the probability of the observation is $P(z_t | x_{q,t})$.

\subsection{Belief Tensor}
Since the state of the robot is initially unknown. Therefore, we assume all free space in the map are possible robot locations and construct a dense belief tensor $\mathcal{B}_t: W \times H \times \Theta$ at time $t$ where $W \times H$ is the width and height of the occupancy map and $\Theta$ is the orientation as shown in Fig~\ref{fig:heatmap}.
A location $[i, j, k]$ in the belief tensor represents a robot state $x_t = \{i \delta_W, j \delta_H, k \delta_{\Theta} + \theta_t\}$ where $\delta_W$, $\delta_H$, and $\delta_{\Theta}$ is the resolution of $\mathcal{B}$ and $k$ starts from zero. $\theta_t$ is the current rotational angle with respect to the initial state. The internal orientation is then updated with
$\theta_t = \theta_{t-1} + w$ given a motion update $u_t = \{u, v, w\}$.


\subsubsection{Motion update}
We update each channel of the tensor independently with the movement $u_t = \{u, v, w\}$. For a channel $k$, the corresponding motion is defined as the following.
\begin{equation}
    u_{t,k} = R_{x_t} [u, v]^T
\end{equation}
where $R_{x_t}$ is the rotation matrix from the orientation in $x_t$.
Therefore, in the first channel, the motion update is just $u_{t,0} = [u, v]$ since $R_{x_0}$ is identity.

The specific motion update for each channel can be done using an affine transformation with matrix.
$$
\begin{bmatrix}
1 & 0 & u_{t, k} \\
0 & 1 & v_{t, k} \\
\end{bmatrix}
$$
The resulting belief tensor is 
\begin{equation}
    \mathcal{B}'_t = \mathcal{AF}(\mathcal{B}_{t-1}, u_t)
\end{equation}
where $\mathcal{AF}(\cdot)$ is the affine transformation function. 

\subsubsection{Motion uncertainty}
For a discrete region $q$, the probability that the robot state is at location $q$ is given as 
\begin{equation}\label{eq:motion_update}
P(x_{q,t} | u_{1:t}) = \sum_{i \in M} P(x_{q,t} | u_t, x_{i,t-1}) P(x_{i,t-1})
\end{equation}
where $u_{1:t}$ is all the observations from time $1$ to $t$.

Using Equation~\ref{eq:motion_update}, we can update the probability of each state in the tensor by aggregating the surrounding values. Assuming our motion uncertainty is a Gaussian distribution with covariance matrix defined as $\Sigma_t \in R^{3 \times 3}$, the Gaussian kernel is the expressed as 
\begin{equation}
    G(x_t, \Sigma_t) = \exp(-\frac{1}{2} (x_t - \overline{x_t})^T \Sigma_t^{-1} (x_t - \overline{x_t}))
\end{equation}
where the covariance matrix is constructed as 
\begin{equation}
    \Sigma_t = R_z(x_t)^T \ diag(\sigma_x, \sigma_y, \sigma_{\theta}) R_z(x_t)
\end{equation}
where $R_z(x_t) \in R^{3\times 3}$ is the rotation along the $z$ axis with the angle from state $x_t$ and $diag(\cdot)$ is the diagonal matrix of a vector.


We can generalize the operation as 
\begin{equation}
    \mathcal{B}''_{t, k} = \mathcal{G}(\mathcal{B}'_t, k)
\end{equation}
where the function $\mathcal{G}(\cdot)$ is the 3D convolution function. Since the orientation in $x_t$ for each channel of the tensor is dependent on $k$, it is then an input to the function $\mathcal{G}(\cdot)$.

As shown in Fig~\ref{fig:heatmap}, each channel is a belief map representing the state probability with different orientation. 

\subsubsection{Map sensor update}
Each time after applying the affine transformation and motion blur, the final "observation" update of odometry-only belief tensor is multiplied with the map probability $\mathcal{M}$. Since $P(m_{occ} \in M) = 0$ for obstacles and $P(m_{free} \in M) = 1$ for free space, the probability in obstacle locations will be reset to zeros. The initial state assumption and the trajectory that leads to the obstacles is then an invalid assumption. 

We show the detailed algorithm in the pseudo-code below, where $\odot$ is pairwise multiplication and $\mathcal{G}(\mathcal{B}_0, k)^{-1}$ is the pixel-wise inverse.
\begin{algorithm}[h]
\label{alg:belief_map}
\caption{Belief tensor update}
\begin{algorithmic} [1]
\REQUIRE $\mathcal{B}_0$
\IF{$|u_{t,k}| >= 1$}
\FOR{$k=0$ to $|\Theta|$}
\STATE $\mathcal{B}'_{t,k}  = \mathcal{AF}(\mathcal{B}_{t-1,k}, u_{t,k})$
\STATE $\mathcal{B}'_{t,k}  = M \odot \mathcal{B}'_{t,k}$
\STATE $\mathcal{B}''_{t,k} = \mathcal{G}(\mathcal{B}'_t, k)$
\STATE $\mathcal{B}''_{t,k} = M \odot \mathcal{B}''_{t,k}$
\STATE $\mathcal{B}_{t,k}   = \mathcal{B}''_{t,k} \odot \mathcal{G}(\mathcal{B}_0, k)^{-1}$
\ENDFOR
\ENDIF
\end{algorithmic}
\end{algorithm}

The sensor update needs to be done not only after the entire process at line 6 but also in between motion update and uncertainty update at line 4. It is to avoid the convolution process to extract nearby state that should be zero if just moved from the free space into the occupied regions. 

When the robot travels near a wall, the boundary state probability will decrease much faster comparing to traveling in open space. It is because the state probability are not corrected by the wall but simply removed instead. It is equivalent to the robot moving along side a cliff instead of a wall, which it may fall into. Therefore, we add line 7 to normalize the convoluted belief map by the sum of kernel activation as follows. 
\begin{equation}
  \mathcal{B}_{t,k}   =\mathcal{B}''_{t,k} \odot \mathcal{G}(\mathcal{B}_0, k)^{-1}
\end{equation}
Since $\mathcal{B}_0$ is a binary tensor with ones in free space, the values from $\mathcal{G}(\mathcal{B}_0, k)$ is just the kernel activation sum for each pixel. 

The belief tensor can be converted into a belief map of the environment by taking the maximum value across the channels for each state. The estimates of the robot state is then the maximum probability location in the belief map. 


\subsection{Observation update}
With the belief tensor of the robot state, it is also possible to update the tensor with observations from other sensors such as LIDAR or camera. 
However, it is expensive to compare measurements at all locations to update the tensor values. Therefore, we sample a subset of possible states and update the corresponding belief value as follows.
\begin{equation}\label{eq:sensor_update}
    P(x_{q,t} | u_{1:t}, z_t) = \eta P(z_t | x_{q,t}) P(x_{q,t} | u_{1:t})
\end{equation}
where $\eta$ is the normalization factor such that $\sum_{q}P(x_{q,t} | u_{1:t}, z_t) = 1$.
Using our previously estimated belief map, we can efficiently select a subset of samples that is representative of the belief map distribution. 

\begin{figure}[t!]
    \centering
    \includegraphics[width=0.5\textwidth]{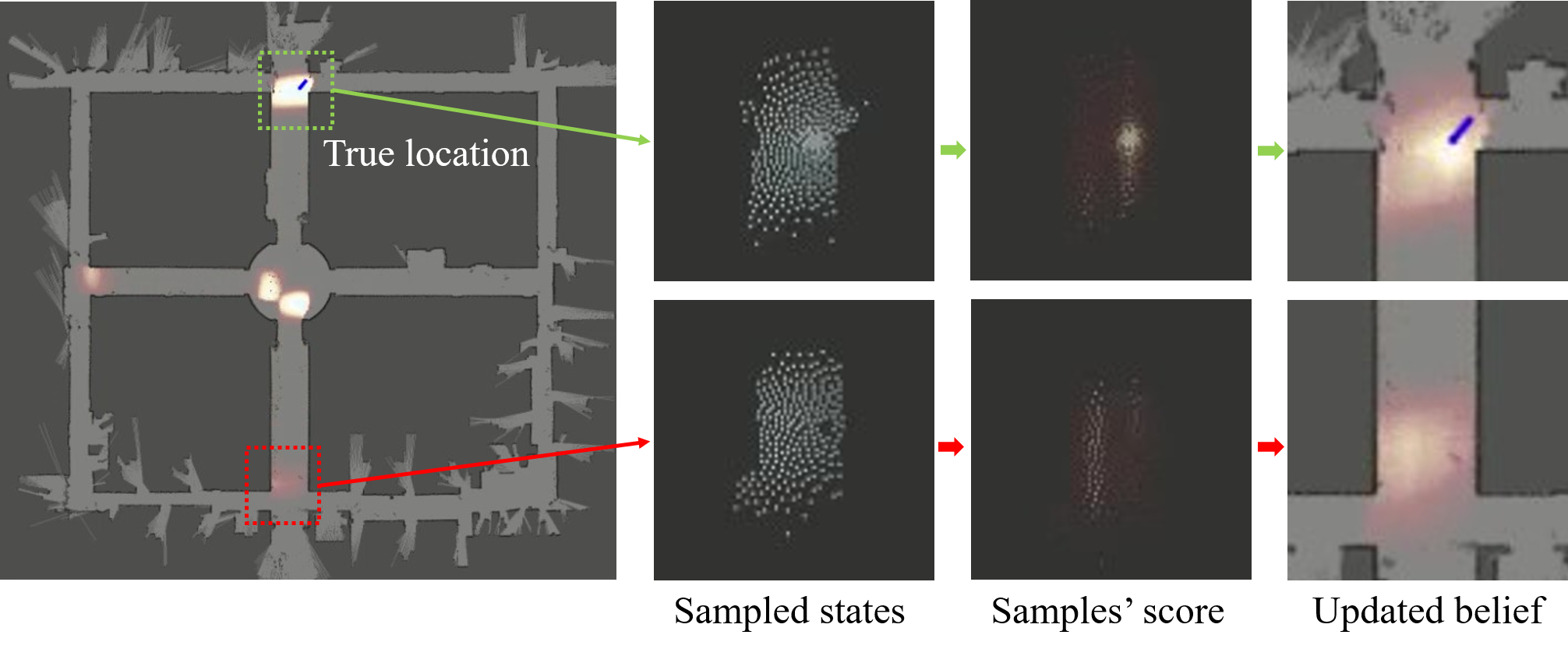}
    \caption{The ground truth location is supported (brighter) by sample updates while the false location diminish (dimmer) with updates.}
    \label{fig:aces_samples}
\end{figure}

\subsubsection{Floyd Steinberg Dithering}
To find a subset of samples that follow the probabilistic distribution of the belief map, we borrow the idea from a image dithering technique call Floyd Steinberg Dithering~\cite{floyd1976}. It is initially introduced as a dithering method to sub-sample the image, which is efficient and simple to implement.
Instead of updating all possible locations in the belief tensor, a subset of points are selected to be updated. 
As shown in Fig~\ref{fig:teaser}(c) and Fig~\ref{fig:aces_samples}, we extract samples from the belief map update the belief tensor using Eq~\ref{eq:sensor_update}.

\subsubsection{Integration with other sensors}
Since the belief tensor is created with only odometry and an occupancy map, with the additional sampling methods, we can integrate observation update with arbitrary sensors such as LIDAR or camera. 
To ensure the localization robustness, the sensor observation model has to satisfy the following equation.
\begin{equation}\label{eq:obs_constraint}
    P(z_t | x^*_t) \geq P(z_t | x_t) 
\end{equation}
where $x^*_t$ is the ground truth location and $x_t \neq x^*_t$ is any other location at time $t$.
It is important to note that symmetry and lack of features have been significant challenges to location detection~\cite{GalvezTRO12, oliva2001modeling, yang2016fast, dube2017segmatch}.
However, our method relaxes the constraint such that the posterior of the ground truth location $P(x^*_t | u_{1:t}, z_t)$ does not disappear in the event of confused observations.
It means that as long as the observation from other similar looking location is not better than that of the correct location, the accuracy is guaranteed. 

\section{Experiments}
\begin{figure}[t!]
\centering
    \begin{subfigure}[b]{0.261\textwidth}
       \includegraphics[width=\textwidth]{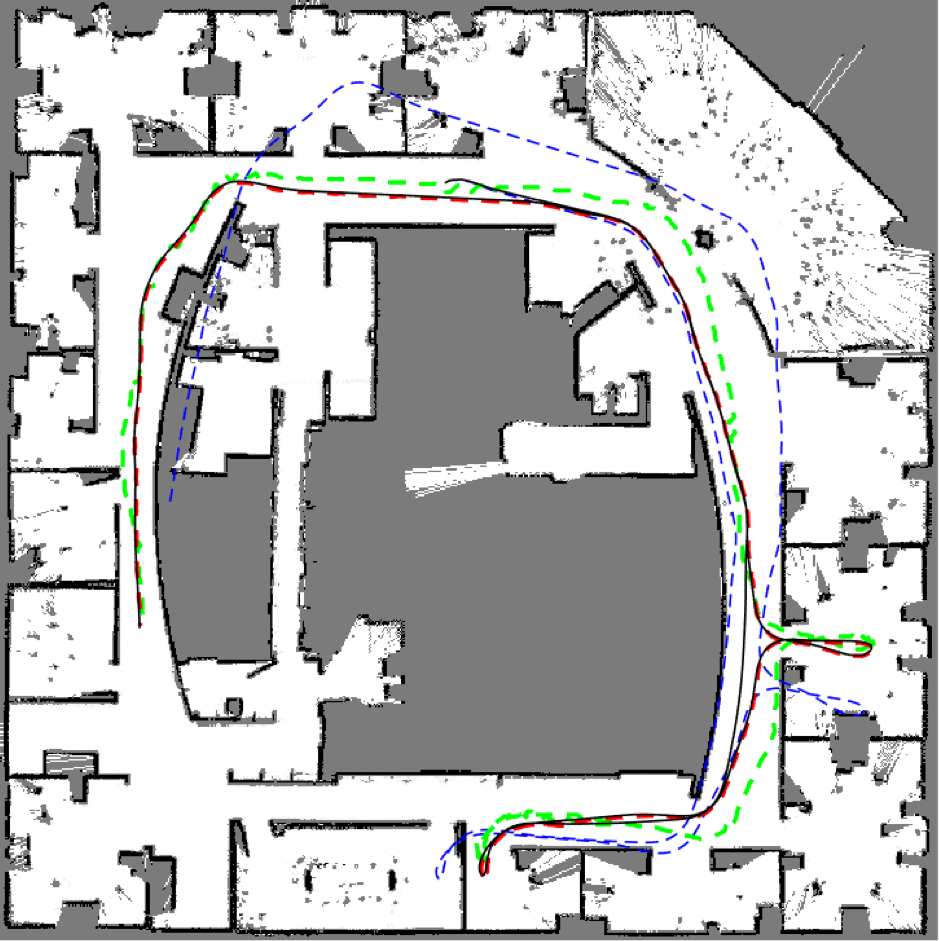}
       \vspace*{-6mm}
       \caption{}
       \vspace*{1mm}
   \end{subfigure}
   \begin{subfigure}[b]{0.19\textwidth}
       \includegraphics[width=\textwidth]{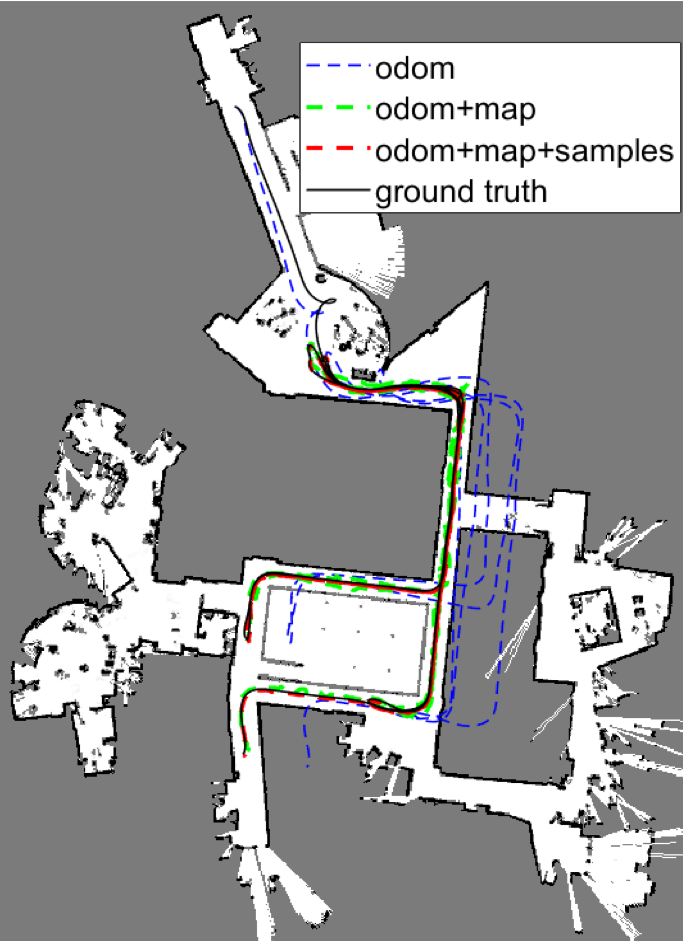}
       \vspace*{-6mm}
       \caption{}
       \vspace*{1mm}
    \end{subfigure}
    \caption{Comparisons among ground truth trajectory to our method with and without sample updates. The odometry will drift over time but odometry + map will stay within the vicinity of the ground truth location. With LIDAR sample updates, the final trajectory overlaps with the ground truth.}
    \label{fig:traj_comp}
\end{figure}

\begin{figure}[t!]
\centering
    \includegraphics[width=0.5\textwidth]{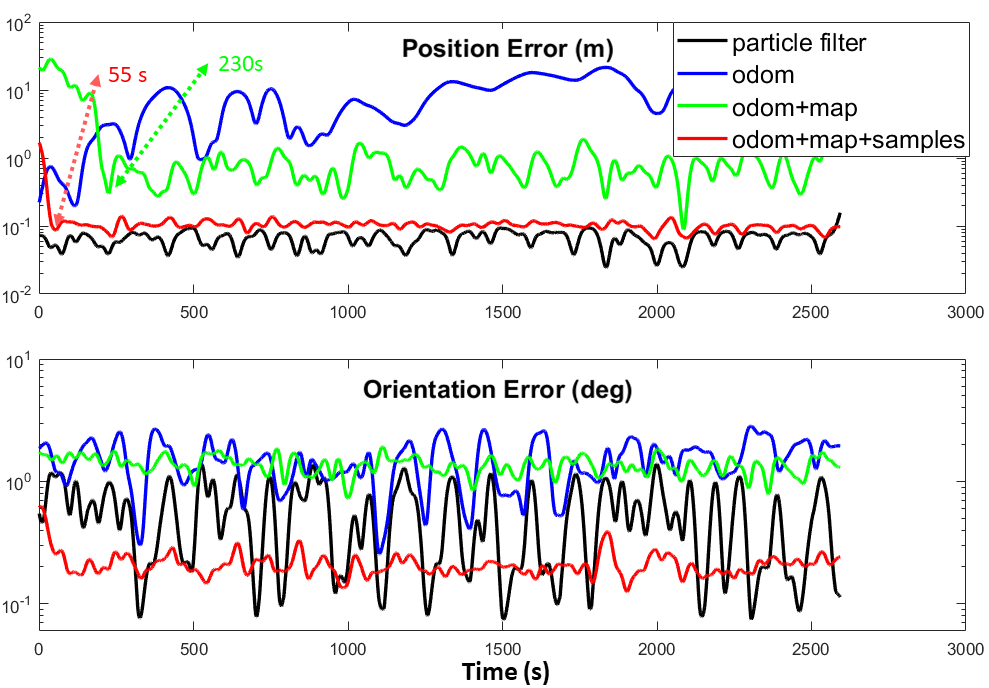}
    \caption{The long-term localization error comparison. The position error converges after 230 seconds for our method w/o sample updates and 55 seconds for our method w/ sample updates.}
    \label{fig:loc_err}
\end{figure}

To evaluate our localization method, we conduct experiments in different environments from a public data set~\cite{Radish}.
The algorithm is written in c++ and runs on an Ubuntu system with i7 Intel core and a Titan V graphic card.
The simulation environment is from ISAAC SDK~\cite{isaacsdk} which provides realistic LIDAR sensor measurements for any given floor plans. 
We compare our algorithms with a particle filter method~\cite{gustafsson2010particle} that starts at the highest $P(z_0 | x_0)$ for all $x_o \in \mathcal{M}$. A total of 75 particles are used to track the local state from the estimated initial location. In this section, we show the robustness of our algorithm, by real experiments, in terms of (1) fast and guaranteed convergence at the location where prior methods have ambiguities, (2) accuracy and stability in long time localization task, and (3) the capability to be fused with other sensors for performance improvement. 
\begin{table*}[t]
    \begin{center}
    \centering
        \begin{tabular}{ |c||c|c|c|c|c| }
             \hline
               & min error (m) & median error (m) & max error (m) & wrongly localized ($ > 1m$) & Dimension (pixel = 0.1m) \\
             \hline
             ACES3 Austin & 15.3 & 44.1 & 132.0 & 8.5\% & $586 \times 556$\\
             \hline
             Belgioioso Castle & 9.9 & 32.6 & 105.3 & 1.2\% & $993 \times 206$ \\
             \hline
             MIT CSAIL & 9.4 & 27.9 & 63.4 & 1.0\% & $482 \times 668$\\
             \hline
             Intel Research Lab & 18.4 & 93.7 & 362.5 & 0.2\% & $579 \times 581$\\
             \hline
             Seattle, UW & 4.9 & 30.8 & 79.2 & 0.7\% & $908 \times 229$\\
             \hline
        \end{tabular}
        \caption{Localization convergences for different maps.}
        \label{tab:localization_convergence}
    \end{center}
\end{table*}

\begin{table*}[t!]
    \centering
    \begin{tabular}{|c|c||c|c|c|c|c|c|}
        \hline
        \multicolumn{2}{|c||}{}& \multicolumn{2}{c|}{ 0 to 1 mins} & \multicolumn{2}{c|}{1 to 5 mins} & \multicolumn{2}{|c|}{5 to 30 mins} \\
        \cline{3-8}
        \multicolumn{2}{|c||}{}& XY error(m) & $\theta$ error(deg)&  XY error(m) & $\theta$ error(deg)& XY error(m) & $\theta$ error(deg)\\
        \hline
        \multirow{3}{*}{ACES3 Austin}
        & particle filter &  \textbf{0.054} & 0.757 & \textbf{0.050} & 0.566 & \textbf{0.062} & 0.677 \\
        & ours w/o samples & 15.339 & 1.471 & 0.613                  & 1.407 & 0.603 & 1.449 \\
        & ours w/ samples & 4.913 & \textbf{0.367} & 0.117           & \textbf{0.204} & 0.106 & \textbf{0.209} \\
        \hline
        \multirow{3}{*}{Belgioioso Castle} 
        & particle filter & \textbf{0.038} & 1.102 & 0.039          & 0.431 & \textbf{0.034} & 0.358 \\
        & ours w/o samples & 11.068 & 1.355 & 14.713                & 1.385 & 0.496 & 1.422 \\
        & ours w/ samples & 1.808 & \textbf{0.648} & \textbf{0.101} & \textbf{0.239} & 0.129 & \textbf{0.204} \\
        \hline
        \multirow{3}{*}{CSAIL MIT}
        & particle filter & \textbf{0.038} & \textbf{1.102} & \textbf{0.039} & 0.431 & \textbf{0.034} & 0.358 \\
        & ours w/o samples & 20.121 & 1.183 & 4.997                          & 1.644 & 0.415 & 1.466 \\
        & ours w/ samples & 1.099 & 1.152 & 0.200                            & \textbf{0.265} & 0.129 & \textbf{0.252} \\
        \hline
        \multirow{3}{*}{Intel Research Lab} 
        & particle filter & \textbf{0.087} & \textbf{0.552} & 0.070 & 0.662 & 0.072 & 0.623 \\
        & ours w/o samples & 10.928 & 1.462 & 0.996 & 1.419 & 0.965 & 1.408 \\
        & ours w samples & 5.325 & 0.723 & \textbf{0.066} & \textbf{0.243} & \textbf{0.065} & \textbf{0.208} \\
        \hline
        \multirow{3}{*}{Seattle, UW} 
        & particle filter & \textbf{0.059} & 1.033 & \textbf{0.062} & 0.591 & \textbf{0.065} & 0.514 \\
        & ours w/o samples & 24.356 & 1.581 & 4.657 & 1.454 & 0.751 & 1.345\\
        & ours w/ samples & 0.548 & \textbf{0.295} & 0.103 & \textbf{0.205} & 0.104 & \textbf{0.206} \\
        \hline
    \end{tabular}
    \caption{Mean long term localization error of each map within different time intervals.}
    \label{tab:long_term_err}
\end{table*}
The particle filter runs at $\sim20$ ms and our method runs at $\sim3.2$ ms on average without sample updates. With sample updates from LIDAR sensor, the average run-time is $\sim31.2$ ms. Our method achieves real-time performance using a belief tensor of size $500\times500\times128$ (depends on the actual map size), with pixel resolution of 0.1 meters and angular resolution of $2.81^{\circ}$ respectively. 

\subsection{Information from the map sensor}
To demonstrate the information collected from the "map sensor", we show the qualitative results in Fig~\ref{fig:traj_comp}. The initial estimates are discarded, since the robot needs to move around (100 meters) to localize itself. As is shown in Fig~\ref{fig:traj_comp}, The odometry trajectory drifts from ground truth trajectory. The trajectory from the map-corrected odometry shows no large divergences and stays in the vicinity of the ground truth trajectory, though not quite accurate. It is expected since any precise location information is not given. After fusing with LIDAR observation, the final trajectory overlays with the ground truth trajectory. Fig~\ref{fig:loc_err} shows the quantitative results where the odometry + map trajectory stays within around 1 meter accuracy to the ground truth after converging. 

\subsection{Localization robustness}
Previous methods do not track dense robot states at all times due to computational limitations. They assume that the initial location is known or is at the region with the highest sensor response, and start tracking following only local information. Since there is only one ground truth, a tracking method that assumes locally distributed noise is used. However, such mechanism may fail in the existence of observation ambiguity due to topology symmetry. 
Fig~\ref{fig:aces_samples} shows the example of narrow corridors where localization methods based on LIDAR observation encounter confusion.

To characterize the localization difficulty in different maps, we evaluate the localization error of LIDAR observations at each possible robot location. More specifically, for each location $p = [x, y]$ with observation $z$, we estimate the location $\hat{p} = [\hat{x}, \hat{y}]$ that best matches $z$. In perfect setting, $p = \hat{p}$, but it is not always true in the existence of noise. The localization error is then defined as $e_p = ||p - \hat{p}||_2$ for each location $p$. Therefore, the localization difficulty of a map can be estimated as 
\begin{equation}
    D_{\mathcal{M}} =\frac{\sum_{p \in \mathcal{M}} I(e_p > C)}{\sum_{p \in \mathcal{M}}p}
\end{equation}
where $I(\cdot)$ is the indicator function that outputs $1$ if $e_p > C$, $C = 1$ meter, and $\sum_{p \in \mathcal{M}}p$ is the number of free space in map $\mathcal{M}$.

The wrongly localized column in Table~\ref{tab:localization_convergence} shows different maps' difficulty level. Maps with higher difficulty are the ones with similar hallways where the LIDAR measurements cannot be distinguished. In the map \textit{ACES3 Austin}, $8.5\%$ wrongly localized regions correspond to more than $28,000$ locations at which the map is ambiguous. Even in the map with the lowest difficulty (\textit{Seattle, UW}), more than 1400 locations can be wrongly localized with at least 1 meter error. 

We show that our method is robust to those ambiguous regions and is able to accurately localize the robot by moving less than 50 meters. The robot starts at 500 arbitrary locations including 250 wrongly localized regions for each map. Since the initial location is unknown, the robot need to travel for a certain distance until localized. The robot's travel policy is random walk with no external guidance. Once the localization error is less than $10$ cm for more than $1$ mins, the robot is considered localized. The distribution of the convergence distance is shown in Table~\ref{tab:localization_convergence}. 
There are a few cases where the convergence distances are large ($> 200$m). They result from the exploration trajectory overlapping itself due to random motion. For example, in the Intel research map shown in Fig~\ref{fig:traj_comp}, the structure tolerates circular robot motion without being able to pin point the exact location. 

\begin{figure}[t]
    \centering
    \includegraphics[width=0.5\textwidth]{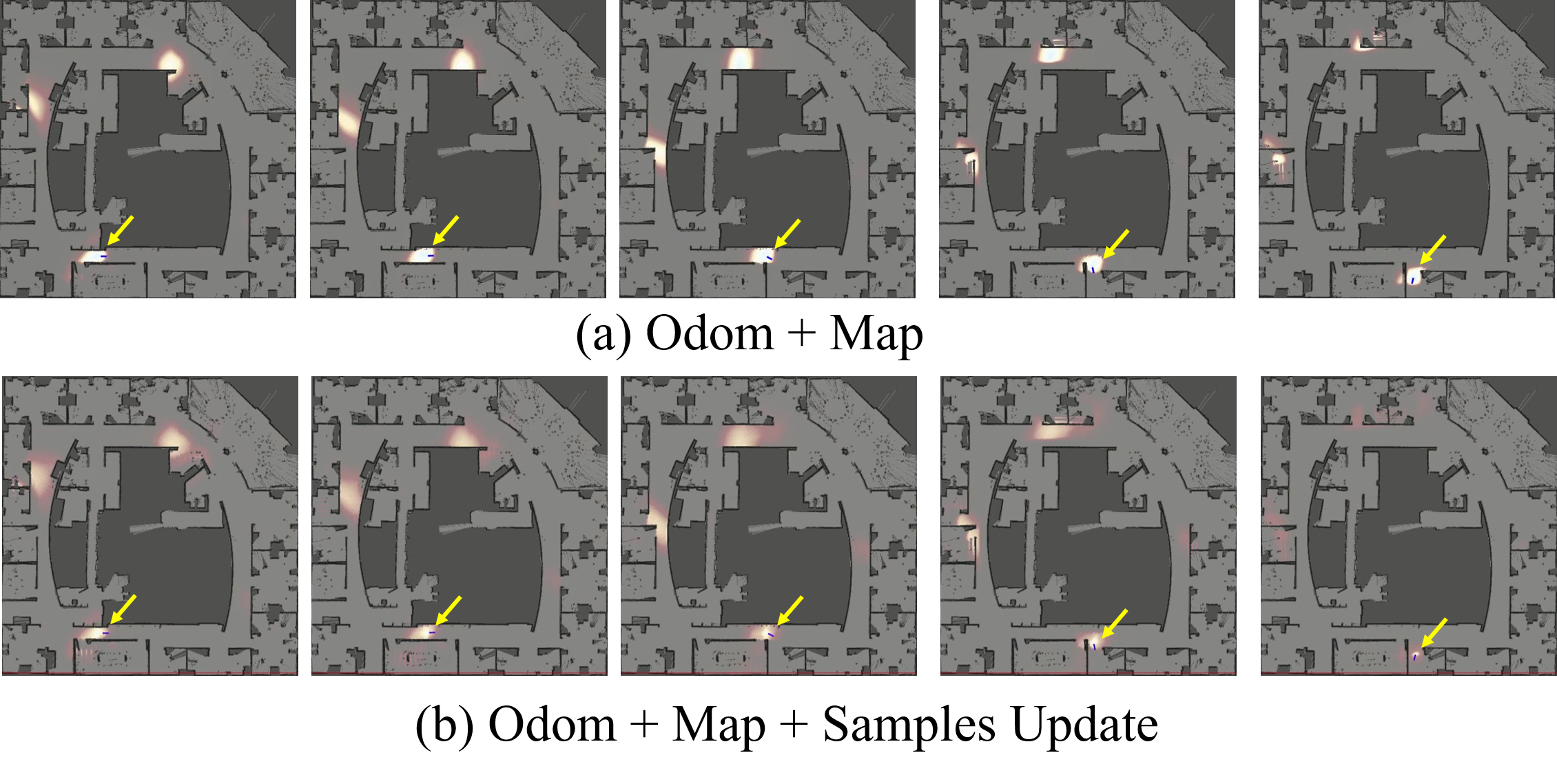}
    \caption{Both (a) and (b) have more than one high probability regions. As more unique structures are observed in (b), the incorrect locations disappear faster than those in (a).}
    \label{fig:intel_expr}
\end{figure}

\begin{figure}[t]
    \centering
    \includegraphics[width=0.5\textwidth]{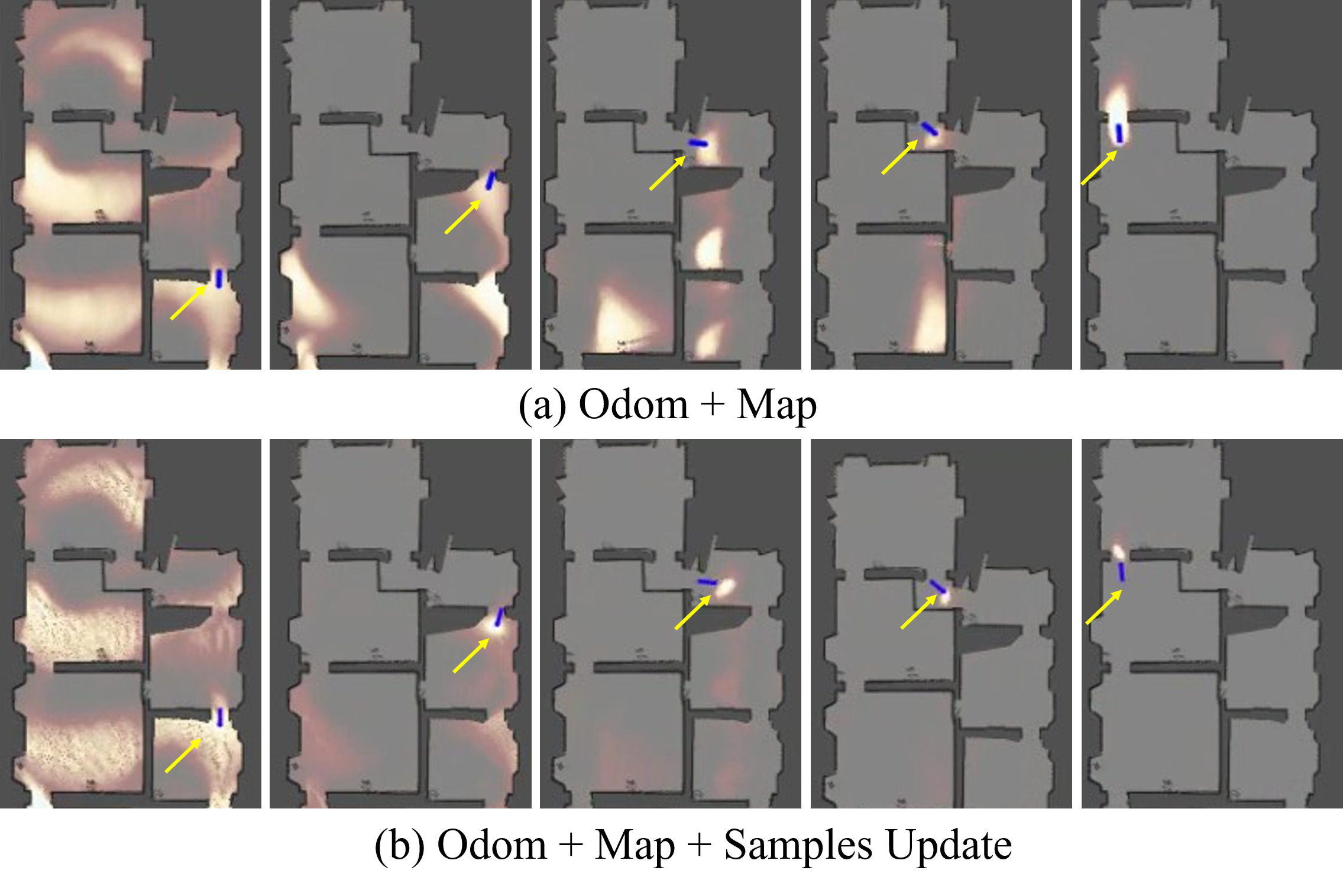}
    \caption{While multiple hypotheses survives in (a) initially, the unique motion pin points region that surrounds the ground truth. (b) converges much faster than that in (a) because of the additional observations from LIDAR measurements.}
    \label{fig:bel_expr}
\end{figure}

When ambiguous observations are encountered, our method determine the exact location by moving a short distance that aggregates the unique observations along the trajectory. 
The belief tensor will always maintain a high probability at the true location as shown in Fig~\ref{fig:bel_expr}. Even though there may exist other incorrect locations, they will not survive for long. Fig~\ref{fig:aces_samples} and Fig~\ref{fig:intel_expr} show the cases where there are multiple high probability estimation of the robot location. However, as additional observations and the traversability constraints are imposed, only the true location survives. When auxiliary sensors are used, we can guarantee that the algorithm will not lose track of the ground truth location as long as the constraint from Eq~\ref{eq:obs_constraint} is satisfied. The capability to differentiate using a sequence of observations is thus an inherent property of our tracking method.


\begin{figure}[h]
    \centering
    \includegraphics[width=0.5\textwidth]{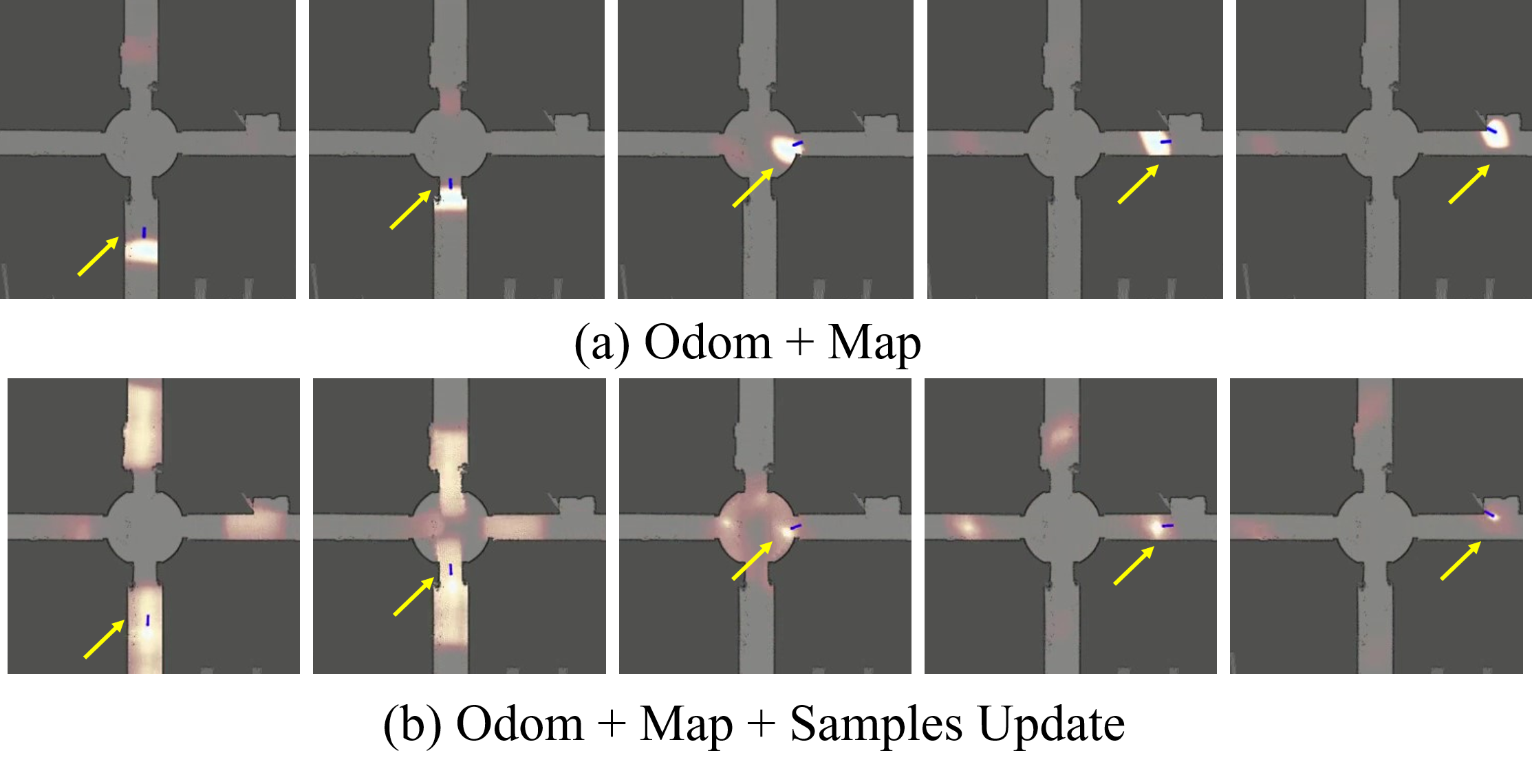}
    \caption{While (a) only have one high probability region, (b) has multiple due to similar LIDAR observations in the hallway. Despite the confusion from LIDAR, as the robot travels to a unique location, the region converges again to the ground truth.}
    \label{fig:aces_expr}
\end{figure}

\subsection{Long term localization accuracy}
We run our algorithms for 30 mins to test the long term localization error in all maps. Table~\ref{tab:long_term_err} shows the accuracy comparison with traditional particle filter. It is shown that our method achieves superior performance on orientation accuracy with competitive performance on location accuracy after initial stage of information integration. 
After 5 mins, our methods without samples can also provide good localization up to 1 meter accuracy. Fig~\ref{fig:loc_err} shows one instance for the long term error comparison. 

\section{Conclusion}
In conclusion, we propose a method to use occupancy map as an additional sensor such that a blind robot with only odometry is able to localize itself in real-time. We also adapt our algorithm to other auxiliary sensors (such as LIDAR) efficiently. The resulting algorithm is able to provide a robust belief tensor of the robot state with accurate localization. Since our method constantly tracks dense robot states, it is robust and can recover from localization failure in the event of ambiguous observations. We validate our algorithm in publicly available floor plans which contains difficult scenarios such as symmetric structures. The experiments show that our method can successfully localize the robot despite the ambiguities. We can also achieve real-time performance (up to $\sim 300$Hz for odometry + map and $\sim 30$Hz with LIDAR samples). 
The convergence distance histogram can be heavily tailed. Therefore, we plan to explore active localization so that the distance traveled until convergence is minimized. 
\section*{Acknowledgement}
The authors of this paper would like to thank Dmitry Chichkov, Benjamin Butin, Qian Lin, and Daniel Abretske who provide tremendous help on debugging and accelerating the algorithm. The authors would also like to thank Shan Su for her supports and valuable discussions.
Thanks go to Cyrill Stachniss, Dirk Haehnel, and Patrick Beeson for providing the map data~\cite{Radish}.


\newpage
\bibliographystyle{splncs}
\bibliography{egbib}


\end{document}